\documentclass{article}
\usepackage{spconf,amsmath,graphicx}
\usepackage{url}

\usepackage{enumitem}
\setlist{nosep, leftmargin=14pt}

\usepackage{microtype}
\usepackage[square, sort, comma, numbers, compress]{natbib}

\begin{document}
% Title.
% ------
\title{Evaluating unsupervised contrastive learning framework for MRI sequences classification}

%
%\vspace{-10pt}
\address{%
$^{1}$ Department of Biomedical Engineering, Johns Hopkins School of Medicine, USA\\
     $^{2}$ Siemens Healthineers, Malvern, PA, USA\\}
%
% \begin{document}
%\ninept
%
\maketitle

\begin{abstract}

\textbf{The automatic identification of Magnetic Resonance Imaging (MRI) sequences can streamline clinical workflows by reducing the time radiologists spend manually sorting and identifying sequences, thereby enabling faster diagnosis and treatment planning for patients. However, the lack of standardization in the parameters of MRI scans poses challenges for automated systems and complicates the generation and utilization of datasets for machine learning research. To address this issue, we propose a system for MRI sequence identification using an unsupervised contrastive deep learning framework. By training a convolutional neural network based on the ResNet-18 architecture, our system classifies nine common MRI sequence types as a 9-class classification problem. The network was trained using an in-house internal dataset and validated on several public datasets, including BraTS, ADNI, Fused Radiology-Pathology Prostate Dataset, the Breast Cancer Dataset (ACRIN), among others, encompassing diverse acquisition protocols and requiring only 2D slices for training. Our system achieves a classification accuracy of over 0.95 across the nine most common MRI sequence types.}

\end{abstract}

%\vspace{-8pt}
\section{Introduction}

Magnetic Resonance Imaging (MRI) provides clinical experts with detailed 3D visualization and analysis of a patient’s organ tissues, enabling the detection of abnormalities \cite{hussain2022modern, wang2023investigation,feng2023label}. MR images are generated by recording the signal intensities emitted by water protons in tissue when excited by a resonant electromagnetic radio-frequency field. This process produces images with visible contrast between tissue elements (e.g., fat, fluids) by exploiting their varying characteristics, such as proton density and relaxation times. Typically, a comprehensive understanding of abnormalities is achieved by analyzing multiple MRI sequences, including T1-weighted, T2-weighted, FLAIR, Diffusion-Weighted Imaging (DWI), and derived Apparent Diffusion Coefficient (ADC) maps, among others.

Certain MRI sequences are often reviewed together by radiologists because they provide complementary diagnostic information \cite{gu2023exploring,toruner2024artificial}. For example, T2-weighted MRI and DWI are frequently used in tandem for diagnosing lymphoma \cite{gu2011whole,wang2023automated}. During the acquisition of these sequences, relevant patient and imaging information is stored in the DICOM header, including fields such as “Body Part Examined," “Procedure Step Description," “Series Description," and “Protocol Name." However, these fields often lack informative descriptions specific to sequence types, and discrepancies may arise. Such discrepancies occur frequently, especially when default scanner protocol information is used, and technologists do not have sufficient time to update every DICOM field accurately during a busy clinical day.

\begin{figure*}
\includegraphics[width=0.8\textwidth]{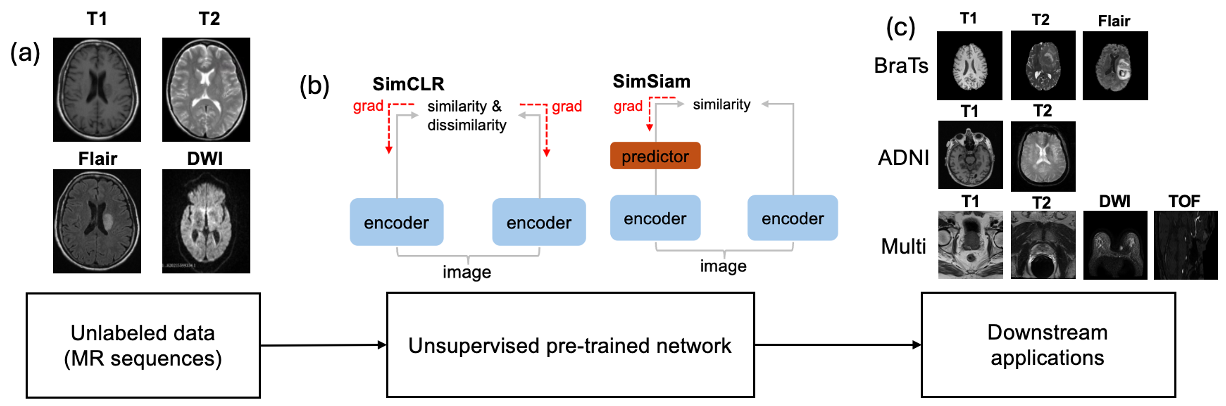}
\centering
%\vspace{-8pt}
\caption{The pipeline of our framework includes (a) unlabeled data from various MRI sequences; (b) an unsupervised pre-training stage using either the SimCLR or SimSiam framework; and (c) a fine-tuning stage applied to downstream tasks on different datasets, including BraTS, ADNI, and the Multi-Organs dataset.} 
%\vspace{-10pt}
\label{fig_workflow}
\end{figure*}

This issue is further compounded by the diversity of MRI protocols and parametric sequences across different institutions globally. Additionally, the use of multiple MRI scanners from various manufacturers introduces extensive variations in voxel intensity distributions across sequences, as illustrated in Fig.~\ref{fig_workflow} (a). Consequently, radiologists and referring physicians must navigate these inconsistencies while performing thorough evaluations of disease status. Such variations can also disrupt radiologists' hanging protocols for reading images in PACS, often necessitating manual intervention to correct discrepancies. These inconsistencies present significant challenges when constructing large-scale clinical cohorts for patients who have undergone specific MRI imaging types or when developing artificial intelligence (AI) algorithms tailored for medical imaging.

An automated method for classifying MRI sequences can significantly reduce the need for radiologists to manually oversee them, enhancing efficiency and accuracy. To address this need, Mello et al., \cite{de2021deep} achieved a classification accuracy of 99.27\% with the ResNet-18 architecture trained on 3D brain MRI volumes. \cite{liang2021magnetic} proposed a method for automatic MRI sequence annotation by training a convolutional neural network (CNN) to classify sequence types, utilizing a 2D perspective to minimize the high memory and computational demands associated with 3D convolutions. However, this approach relies heavily on DICOM header data for accurate categorization and supervision. Moreover, achieving high classification accuracy and model generalizability demands a large quantity of labeled data, which may not always be readily available and requires time-consuming manual verification to ensure label accuracy.

To date, there is no existing literature on unsupervised methods for MRI sequence classification applied to different anatomical regions such as the brain, chest, and abdomen. This paper aims to bridge this gap by proposing an automated framework for the classification of multi-parametric MRI sequences acquired at the chest, abdomen, and pelvis levels, as detailed in Fig.~\ref{fig_workflow}. The proposed algorithm distinguishes between different MRI sequence types, specifically T1w, T2w, Fluid-Attenuated Inversion Recovery (FLAIR), Time-of-Flight (TOF), Trace-Weighted Imaging (TraceW), Diffusion-Weighted Imaging (DWI), ADC, Gradient Echo Imaging (GRE), and Perfusion. We validate our model's effectiveness through consistent accuracy (ACC) metrics across two unsupervised contrastive learning architectures: SimCLR and SimSiam, applied to MRI sequences of both the body organs and brain. 

The structure of this paper is as follows: Section 2 outlines the dataset curation and the proposed method, while Section 3 details the experimental methodology and results. Section 4 provides the conclusions and discusses future perspectives.

%\vspace{-7pt}
\section{Dataset and Methods}
\subsection{Dataset Curation}\label{dataset}

For the in-house internal dataset, a total of 72,188 Brain MRI studies from over 25 distinct sequences were collected, with each series within a study manually verified. The majority of patients were imaged using the nine most common specific MRI sequences: T1, T2, FLAIR, TOF, TraceW, DWI, ADC, GRE, and Perfusion. To create the final cohort, we randomly selected images from 220 studies per sequence, retaining only the central 30\% of images in the sagittal, coronal, and axial planes. Empty sequences contained errors or displayed repeated regions were excluded. Data from the 1,980 patients were then randomly partitioned at the patient level into training (70\%, 864 studies or 23,568 images), validation (10\%, 123 studies or 6,843 images), and test (20\%, 247 studies or 3,568 images) sets, ensuring that all studies from the same study remained in the same subset. To reduce computational load during training, each MRI series was resampled to 84 × 84 pixels.

To assess the model’s performance and generalizability across various anatomical regions and MRI sequences, we also incorporated two publicly available brain image datasets: the Brain Tumor Segmentation (BraTS) dataset \cite{menze2014multimodal} and the Alzheimer’s Disease Neuroimaging Initiative (ADNI) dataset \cite{petersen2010alzheimer}. Additionally, we utilized three organ-specific datasets, including the Fused Radiology-Pathology Prostate Dataset \cite{madabhushi2016fused}, the Lausanne TOF Aneurysm Cohort \cite{ds003949}, and the Breast Cancer Dataset (ACRIN) \cite{newitt2021acrin}. These images were uniformly resampled to 80 × 80 pixels. Given the variation in available sequence types across these datasets, the model was retrained for each dataset separately.

%\vspace{-5pt}
\subsection{Pre-training: Unsupervised Contrastive Learning}

Contrastive self-supervised methods assume that image transformations do not change an image’s semantic meaning. Consequently, different augmentations (including flip, rotation, and elastic transformations in this paper) of the same image form a "positive pair," while other images and their augmentations are "negative pairs" relative to the instance. The model is trained to minimize the distance between positive pairs in latent space while maximizing the separation from negative pairs, using various distance metrics within the contrastive loss function. SimCLR \cite{chen2020simple} showcased the effectiveness of this approach, outperforming supervised models on the ImageNet benchmark with 100 times fewer labels. However, SimCLR's reliance on very large batch sizes for optimal performance can be computationally demanding. To address this, SimSiam \cite{chen2021exploring} utilizes siamese networks without requiring negative samples, enabling it to reduce batch size requirements while learning meaningful representations and preserving model performance. In this work, we compared SimCLR and SimSiam during the pre-training stages.

\subsection{Fine-tuning: Supervised Learning}\label{Fine_tuning}

The curated in-house internal training dataset from Section \ref{dataset} was used for pre-training, with the model subsequently fine-tuned in a supervised manner using varying portions (0.5\% to 100\%) of the dataset. Specifically, the in-house internal testing dataset includes all nine MRI sequences. Additionally, we constructed three external datasets for model fine-tuning and evaluation, extracting 220 patient images per MRI sequence from each. The BraTS dataset contains T1w, T2w, and FLAIR sequences; the ADNI dataset includes T1w and T2w sequences; and the Organs dataset comprises T1w and T2w sequences from the Fused Radiology-Pathology Prostate Dataset, DWI sequences from the ACRIN dataset, and TOF sequences from the Lausanne TOF Aneurysm Cohort. For all other MRI sequences, the originals from the in-house internal dataset were used for fine-tuning and testing.

\begin{figure}
\includegraphics[width=0.43\textwidth]{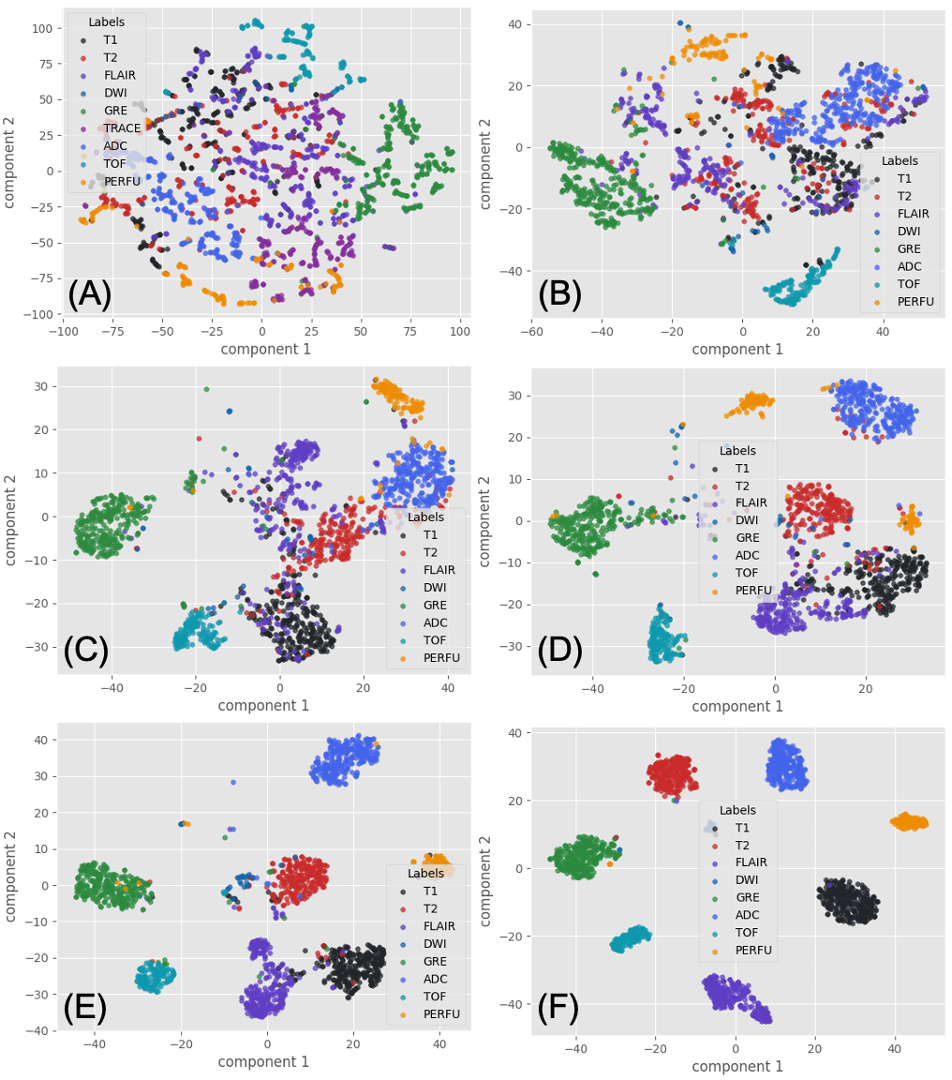}
\centering
%\vspace{-10pt}
\caption{Example of latent space visualizations from SimCLR during pre-training: (a) at epoch 0, (b) at epoch 50. Additionally, visualizations are shown after pre-training and supervised fine-tuning using different percentages of labeled data: (c) 0.5\%, (d) 1\%, (e) 5\%, and (f) 50\%.} 
%\vspace{-10pt}
\label{fig_latent_space}
\end{figure}

%\vspace{-10pt}
\section{Experiments and Results}
We implemented the SimCLR and SimSiam frameworks using ResNet-18 as the backbone for the pre-training phase. To maintain the intensity distribution of the images, we applied only flip, rotation, and elastic transformations for augmentation. Given computational constraints and the convergence behavior of the training and calibration loss, each pre-training session was limited to 50 epochs. Figure~\ref{fig_latent_space} shows latent space visualizations at epochs 0 (Fig.2 (A)) and 50 (Fig.2 (B)). A clear clustering of points is observed after 50 epochs, indicating improved feature representation.

\subsection{Downstream Classification: In-house Dataset}

Using ResNet-18 with unsupervised pre-training weights, we fine-tuned the model’s classification performance in a supervised manner on an in-house proprietary brain image dataset for both the SimCLR and SimSiam frameworks. The latent space visualizations are presented in Fig.\ref{fig_latent_space} (c) to (f). We acknowledge that batch size significantly influences the effectiveness of the unsupervised contrastive loss and, subsequently, the pre-training performance. Therefore, we investigated the optimal batch size for both frameworks in the medical imaging context, testing a range from 64 to 2048. The results are shown in Table 1 and Table 2. Our results indicate that the optimal batch size is 256 for SimSiam and 1024 for SimCLR across most fine-tuning settings using varying proportions of the dataset. The highest classification accuracy for SimSiam was achieved with 50\% of the data and a batch size of 256, while for SimCLR, the highest accuracy was observed with a batch size of 1024 using either 50\% or 100\% of the data for fine-tuning. Additionally, to investigate the effect of resolution on classification performance in SimSiam, we tested different resolutions (64 and 256) while keeping the batch size constant. The summary of the results is shown in Table \ref{simsiam_resolution}, We observed that increasing the resolution to 256 improved classification accuracy compared to a resolution of 64.

\begin{table}
    \centering
    \renewcommand{\arraystretch}{0.9}
    \begin{tabular}{|c|c|c|c|c|c|c|}
        \hline
        {\footnotesize SimSiam} & \textbf{64} & \textbf{128} & \textbf{256} & \textbf{512} & \textbf{1024} & \textbf{2048} \\ \hline
         0.5\% & 0.655 & 0.660 & \textbf{0.704} & 0.657 & 0.625 & 0.619 \\ \hline
        1\% & 0.789 & 0.817 & \textbf{0.827} & 0.780 & 0.803 & 0.780 \\ \hline
         5\% & 0.921 & 0.914 & \textbf{0.920} & 0.910 & 0.925 & 0.915 \\ \hline
        50\% & 0.964 & 0.958 & \textbf{0.967} & 0.959 & 0.962 & 0.966 \\ \hline
         100\% & \textbf{0.963} & 0.959 & 0.959 & 0.9634 & 0.963 & 0.961 \\ \hline
    \end{tabular}
    \label{simsiam}
    %\vspace{-10pt}
    \caption{MR sequence classification performance of the SimSiam model across varying fine-tuning percentages and batch sizes.}
    %\vspace{-10pt}
\end{table}

\begin{table}
    \centering
    \renewcommand{\arraystretch}{0.9}
    \begin{tabular}{|c|c|c|c|c|c|c|}
        \hline
        {\footnotesize SimCLR} & \textbf{64} & \textbf{128} & \textbf{256} & \textbf{512} & \textbf{1024} & \textbf{2048} \\ \hline
        0.5\% & 0.712 & 0.753 & 0.790 & 0.785 & \textbf{0.806} & 0.794 \\ \hline
        1\% & 0.845 & 0.876 & \textbf{0.905} & 0.8884 & 0.879 & 0.875 \\ \hline
        5\% & 0.901 & 0.931 & 0.937 & 0.938 & \textbf{0.939} & 0.930 \\ \hline
        50\% & 0.941 & 0.967 & 0.963 & 0.966 & \textbf{0.968} & 0.961 \\ \hline
        100\% & 0.961 & 0.962 & 0.963 & \textbf{0.968} & 0.967 & 0.962 \\ \hline
    \end{tabular}
    \label{simclr}
    %\vspace{-8pt}
    \caption{MR sequence classification performance of the SimCLR model across varying fine-tuning percentages and batch sizes.}
    %\vspace{-10pt}
\end{table}

\begin{table}
    \centering
    \renewcommand{\arraystretch}{0.9}
    \begin{tabular}{|c|c|c|c|c|}
        \hline
        Simsiam & \textbf{64\_84} & \textbf{64\_256} & \textbf{128\_84} & \textbf{128\_256} \\ \hline
        0.5\% & \textbf{0.674} & 0.438 & 0.570 & \textbf{0.790} \\ \hline
        1\% & 0.829 & \textbf{0.836} & 0.818 & 0.827 \\ \hline
        5\% & 0.909 & \textbf{0.930} & 0.909 & \textbf{0.936} \\ \hline
        50\% & 0.965 & \textbf{0.975} & 0.957 & \textbf{0.971} \\ \hline
        100\% & 0.961 & \textbf{0.968} & 0.958 & 0.964 \\ \hline
    \end{tabular}
    \label{simsiam_resolution}
    %\vspace{-6pt}
    \caption{MR sequence classification performance of the SimCLR model across varying image resolutions. Each column header represents the format \textit{batch size\_resolution}, where, for example, 64\_84 indicates a batch size of 64 and an image resolution of 84 $\times$ 84.}
    %\vspace{-10pt}
\end{table}

%\vspace{-4pt}
\subsection{Downstream Classification: External Datasets}
\begin{figure}
\includegraphics[width=0.46\textwidth]{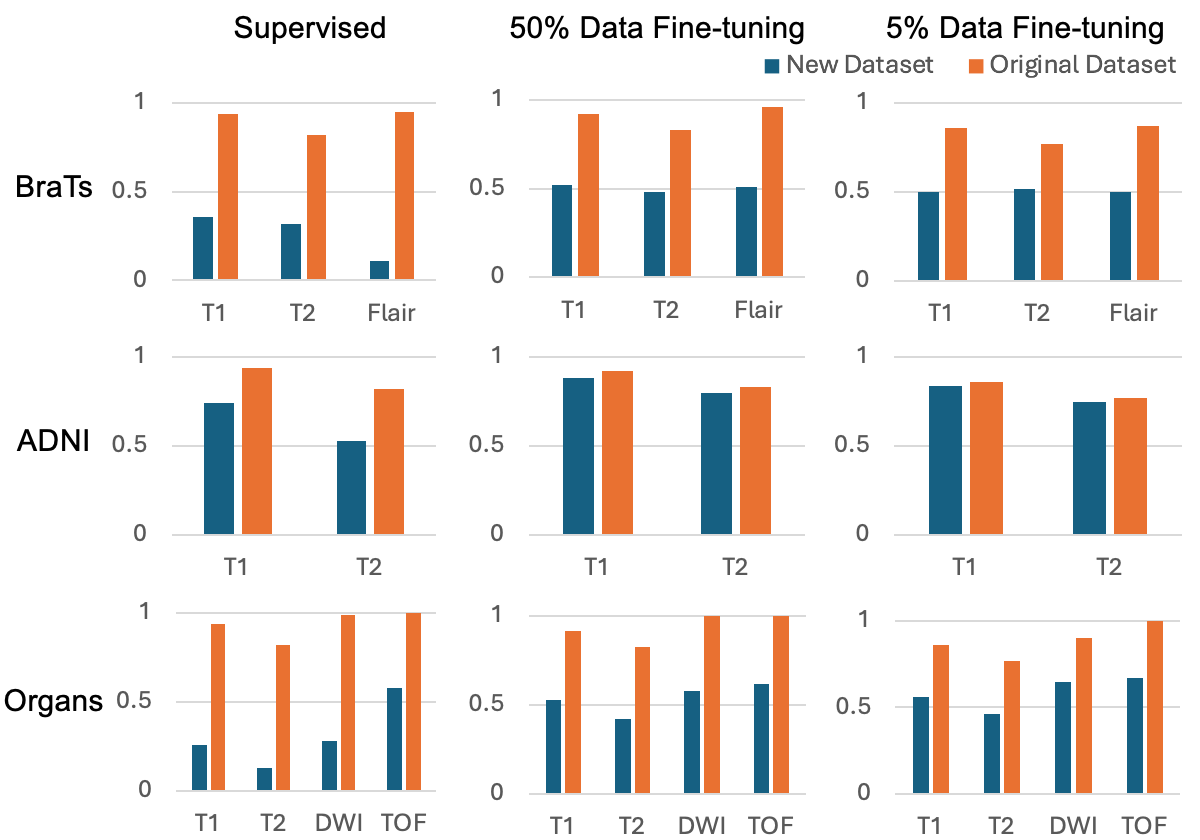}
\centering
%\vspace{-10pt}
\caption{MR sequence classification classification Performance of model on external datasets using varying percentage data for fine-tuning.} 
%\vspace{-15pt}
\label{fig_different_dataset}
\end{figure}

As described in Section \ref{dataset}, we developed three external datasets, including, the BraTS, ADNI dataset, and Organs dataset, specifically for model fine-tuning and testing, extracting 220 patient images per MRI sequence from each dataset. The classification accuracy results for the new MRI sequences from the constructed datasets are shown in Fig.~\ref{fig_different_dataset}. As depicted, a noticeable drop in accuracy is observed across three scenarios: (1) fully supervised training, (2) unsupervised pre-training on the original dataset followed by fine-tuning on the original dataset, and (3) unsupervised pre-training on the original dataset followed by fine-tuning on the new datasets. However, the unsupervised method still demonstrates significantly higher classification accuracy than the supervised approach, likely due to the improved generalizability achieved through unsupervised pre-training.

%\vspace{-15pt}
\section{Discussion and Conclusion}
%\vspace{-8pt}
Both SimCLR and SimSiam effectively classify MRI sequences, showing strong performance across various anatomical regions and modalities. These frameworks demonstrate the potential to improve model generalizability while reducing reliance on extensive labeled datasets, which is crucial in medical imaging.

We applied various data augmentation and downsizing techniques to balance performance and computational efficiency. However, further work is needed to optimize these frameworks, including exploring advanced augmentations, higher-resolution images, and additional MRI sequences. Future studies should also evaluate performance on larger, diverse datasets and clinical scenarios to enhance accuracy and applicability in medical imaging.
%\section{Acknowledgments}
%
%\label{s:acknowledgments}
%

%\bibliographystyle{IEEEbib}
%\vspace{-10pt}
\bibliographystyle{apalike}

\bibliography{unsupervised}

\end{document}